\documentclass[twocolumn,tighten,iop,numbers]{openjournal}
\usepackage{graphicx,amsmath,amssymb,amstext}
\usepackage{amsbsy,amsfonts,amsthm,color}
\usepackage{cancel}
\usepackage{xcolor}
\usepackage[colorlinks,linkcolor=blue,citecolor=blue,urlcolor=blue]{hyperref}
\usepackage[title,toc,page]{appendix}
\let\appendixpagenameorig\appendixpagename
\renewcommand{\appendixpagename}{\centering\textmd{\normalsize\MakeUppercase{\appendixpagenameorig}}}

\usepackage{booktabs}
\usepackage{enumitem}

\makeatletter
\long\def\@makecaption#1#2{%
  \vskip\abovecaptionskip
  \sbox\@tempboxa{#1. #2}%
  \ifdim \wd\@tempboxa >\hsize
    #1. #2\par
  \else
    \global \@minipagefalse
    \hb@xt@\hsize{\box\@tempboxa\hfil}%
  \fi
  \vskip\belowcaptionskip
}
\makeatother

\makeatletter
\long\def\@makecaption#1#2{%
  \vskip\abovecaptionskip
  #1. #2\par
  \vskip\belowcaptionskip
}
\makeatother

\usepackage{tikz}
\usetikzlibrary{shapes,arrows,shadows,fit}
\usetikzlibrary{positioning}
\usetikzlibrary{bayesnet}
\definecolor{ourgreen}{HTML}{06D6A0}
\definecolor{ourblue}{HTML}{118AB2}
\definecolor{darkturquoise}{rgb}{0.0, 0.81, 0.82}
\definecolor{periwinkle}{rgb}{0.8, 0.8, 1.0}

\usepackage{hyperref}
\hypersetup{colorlinks,linkcolor=ourblue,citecolor=ourblue,urlcolor=ourblue}

\graphicspath{{./}{figures/}}

\newcommand{\dif}{\mathop{}\!\mathrm{d}}

\definecolor{ork}{rgb}{0.9,0.1,0.3}
\definecolor{grbl}{rgb}{0.3,0.6,0.7}
\definecolor{bleu}{rgb}{0,0.5,0.6}

\usepackage{listings}
\usepackage{xcolor}

\definecolor{codegreen}{rgb}{0,0.6,0}
\definecolor{codegray}{rgb}{0.5,0.5,0.5}
\definecolor{codepurple}{rgb}{0.58,0,0.82}
\definecolor{backcolour}{rgb}{0.95,0.95,0.92}

\lstdefinestyle{mystyle}{
  commentstyle=\color{codegreen},
  keywordstyle=\color{magenta},
  numberstyle=\tiny\color{codegray},
  stringstyle=\color{codepurple},
  basicstyle=\ttfamily\scriptsize,
  breakatwhitespace=false,         
  breaklines=true,                 
  captionpos=b,                    
  keepspaces=true,                 
  numbers=left,                    
  numbersep=5pt,                  
  showspaces=false,                
  showstringspaces=false,
  showtabs=false,                  
  tabsize=2,
  postbreak=\mbox{\textcolor{red}{$\hookrightarrow$}\space},
}
\begin{document}

\journalinfo{The Open Journal of Astrophysics}
\shorttitle{High-dimensional MCMC Convergence with NF}
\title{\texttt{NFD\lowercase{iag}}: High-dimensional MCMC Convergence Diagnosis with Normalizing Flows\vspace{-4.5em}}

\author{Kutay Nazli$^{1,2}$, Elena Sellentin$^{2,1}$}
\affiliation{$^1$ Leiden Observatory, Leiden University, Einsteinweg 55 2333 CC Leiden, Netherlands}
\affiliation{$^2$ Mathematical Institute, Leiden University, Einsteinweg 55 2333 CC Leiden, Netherlands}
\email{nazli@strw.leidenuniv.nl}
\begin{abstract}
Accurately assessing the convergence of Monte Carlo Markov Chains is crucial for ensuring their validity in Bayesian inference problems. However, convergence assessment in high dimensions is known to be difficult, due to most convergence diagnostics becoming insensitive beyond a few dozen parameters.   Conventional convergence statistics analyze marginal distributions which in high dimensions neglects increasing levels of information from the joint and conditional probability distributions. Thus we propose $\texttt{NFDiag}$, which taps into the information of the full multivariate joint posterior with a normalizing flow (NF). Our convergence diagnostic thereby also analyzes information from conditional distributions. In other words, by working in the full multivariate space, it is sensitive with respect to the geometry of the chain in high dimensions. Our diagnostic utilizes that NFs assume independent and identically distributed (iid) training samples, which is a testable assumption that will be violated by a non-converged or strongly correlated chain. Our diagnostic trains NFs on separate parts of a chain, and then assesses whether these correspond to the same multivariate target distribution. We test our diagnostic in up to 250 dimensions, and demonstrate it is universal and unbiased, and detects problematic chain quality also when Gelman-Rubin $\hat{R}$ and the multivarite effective sample size do not.
\end{abstract}

\maketitle
\section{Introduction} \label{sec:intro}

Bayesian statistics \citep{bayes, bayes?} provides a practical framework for solving scientific inference problems. While some Bayesian inference problems can be  solved completely analytically, most rely heavily on computational methods \citep{gelman_book}, such as Markov Chain Monte Carlo (MCMC) \citep{Metropolis, Hastings} algorithms. Following the initial Metropolis-Hastings algorithm, more variations of MCMC were developed, such as the Gibbs sampler \citep{gibbs1, gibbs2}, the Hamiltonian Monte Carlo (HMC) sampler \citep{hmc}, piecewise deterministic samplers \citep{Joris}, the No U-Turn sampler (NUTS) \citep{nuts} and  the affine invariant sampler \citep{affine_invariant}. 

As these algorithms allowed for sampling from more complex and increasingly higher dimensional probability distributions, a typical research question became: How can one tell that a chain is of sufficient quality to faithfully represent the target distribution? This is usually abbreviated to the question of whether the chain has converged. 

In dimensions beyond a few dozen, this question becomes increasingly hard to answer, and this paper hence proposes a diagnostic for chain quality assessment in high dimensions.

One prerequisite for convergence is that the chain has sufficiently mixed  and explored all the subvolumes of the probability distribution. This is implicitly a question about the geometric shape of the chain in high dimensions. This geometry contains valuable information about whether the chain correctly resolves the conditionals of the target distribution. 

The most popular convergence diagnostics analyze marginals of the chain only. These include  visual inspection of marginal scatter plots, the correlation length of the chain and the effective sample size (ESS) \citep{ess}, Gelman-Rubin's $\hat{R}$ \citep{rhat} and Hanson's diagnostic \citep{hansons}). 

Restricting the analysis to marginals only loses information. For example, unexplored sub-volumes typically become invisible when projecting down onto a 2-dimensional marginal, and a convoluted trace of the chain in the full volume can look like a well-mixed chain in a 2-dimensional projection. In the following, we therefore train normalizing flows (NFs) to use the information in the conditional distributions on top of that in the marginals.

We utilize NFs to construct the posterior of samples from an MCMC chain. NFs assume the samples used to train its neural networks to be independent and identically distributed (iid), and a chain that is yet to converge contradicts this assumption. Thus a NF model built with samples falsely assuming iid will fail to accurately represent another chain from the same target distribution. Our code $\texttt{NFDiag}$ harnesses this fact to render the lack of convergence in MCMC chains visible. It computes a new convergence statistic in high dimensions we have called $\texttt{NFD}$, given by Eq.~\ref{eq:nfd}, which is the main result of this paper.

We structure this work as follows. Further details about NF and the theoretical formulation that makes $\texttt{NFDiag}$ possible are given in Sec.~\ref{sec:methodology} and \ref{subsec:test_intro}. We provide a brief user guide to \texttt{NFDiag} in Sec.~\ref{sec:implementation}. Sec.~\ref{sec:demos} demonstrates our method is universal and unbiased in up to 125 dimensions, and highlights its advantages in comparison with other common statistics. In Sec.~\ref{subsec:science}, we test it on a presently relevant cosmological inference problem in 250 dimensions. In Sec.~\ref{sec:conclusions}, we conclude. 
\section{Methodology} \label{sec:methodology}
\texttt{NFDiag} relies on normalizing flows (NFs) to model the joint probability distribution of a (or part of a) chain of samples acquired from a Markov Chain Monte Carlo (MCMC) sampler. We provide a brief introduction to NFs in Sec.~\ref{subsec:nf_intro}. Crucially, we outline the capabilities of NF models and how they relate to \texttt{NFDiag} in \ref{subsec:sens}.

\subsection{Normalizing Flows} \label{subsec:nf_intro}
NFs aim to represent an intractable probability distribution as a series of variable transformations of a known probability distribution. We denote the former as $P_\mathcal{X}(\textbf{x})$ for $\textbf{x}\in\mathcal{X}$, and the latter with $P_\mathcal{U}(\textbf{u})$ for $\textbf{u}\in\mathcal{U}$. Then, let $\textbf{f}_{1 \leq i\leq n}\in \mathcal{F}$ be $n$ functions in the family of diffeomorphisms $\mathcal{F}$. NFs aim to establish a probability-preserving relation of the form
\begin{equation}
    \textbf{x} = \textbf{f}_n \circ \textbf{f}_{n-1} \circ \ldots \circ \textbf{f}_1(\textbf{u}),
    \label{eq:2.1}
\end{equation}

\noindent between the domains of $P_\mathcal{X}(\textbf{x})$ and $P_\mathcal{U}(\textbf{u})$. Preservation of probability density requires that 
\begin{equation}
    P_\mathcal{X}(\textbf{x}) = \prod_{i}^{n} \left| \text{det}\frac{\dif \textbf{f}_i(\textbf{z}_{i-1})}{\dif\textbf{z}_i} \right| P_\mathcal{U}(\textbf{u}),
    \label{eq:2.2}
\end{equation}

\noindent where $\textbf{z}_i = \textbf{f}_{i}(\textbf{z}_{i-1})$ are representations of the intermediate variables of the transformations (with $\textbf{z}_0 = \textbf{u}$ and $\textbf{z}_{n+1} = \textbf{x}$). The change of variables formula given by Eq.~\ref{eq:2.2} allows the probability of any $\textbf{x}$ to be uniquely calculated without $P_{\mathcal{X}}$, relying only on the Jacobian matrix and the probability $P_\mathcal{U}(\textbf{u})$. The complete mathematical derivation of NF is presented in \cite{flows}.

While $n \gg 1$ is required for above to hold, in practice, it might be impossible to find $\textbf{f}_{i}$ with a finite or small $n$ such that Eq.~\ref{eq:2.2} remains computable in a reasonable amount of time. Thus realistically, Eqs.~\ref{eq:2.1} and~\ref{eq:2.2} are only approximately satisfied, with $\textbf{f}_i$ specifically selected from families of functions with high expressive flexibility. Let $\pmb{\theta}$ denote the hyperparameters of such a set of $\textbf{f}_i$, and $F_{\pmb{\theta}}$ the resulting approximate representation of $P_\mathcal{X}$. $\pmb{\theta}$ are then optimized via minimizing a distance metric between $F_{\pmb{\theta}}$ and $P_\mathcal{X}$ to ensure the approximation is as close to equality as possible.

Conventionally for NF models, the hyperparameters $\pmb{\theta}$ are represented as neural networks (NN), and the distance metric, or loss function, is the forward Kullback-Leibler (KL) divergence \citep{kld}, given by
\begin{equation}
    \begin{aligned}
        D_{\text{KL}}(P_\mathcal{X}||F_{\pmb{\theta}}) &= \int P_\mathcal{X}(\textbf{x}) \text{log} \left( \frac{P_\mathcal{X}(\textbf{x})}{F_{\pmb{\theta}}(\textbf{x})} \right) \dif \textbf{x} \\
        &= -\mathbb{E}_{P_\mathcal{X}(\textbf{x})}[\text{log}F_{\pmb{\theta}}] + \text{constant}.
    \end{aligned}
    \label{eq:2.3}
\end{equation}

Here the constant term is the expectation under $P_{\mathcal{X}}$ of $\log P_{\mathcal{X}}$; since it does not vary as the NF model is trained, we drop it when minimizing $D_{\text{KL}}$. 

\subsection{Sensitivity to Sample Quality} \label{subsec:sens}
In practice, the integral of Eq.~(\ref{eq:2.3}) has to be calculated with a sample-based Monte Carlo estimator. The expectation value in the KL divergence of Eq.~(\ref{eq:2.3}) is approximated by the Monte Carlo integral
\begin{equation}
    \begin{aligned}
        D_{\text{KL}}(P_\mathcal{X}||F_{\pmb{\theta}}) &= -\frac{1}{N} \sum_{n=1}^N \text{log}F_{\pmb{\theta}}(\textbf{x}_n)\\
        &\equiv \mathcal{L}(\pmb{\theta}), 
    \end{aligned}
    \label{eq:2.4}
\end{equation}

\noindent which assumes that the set of samples $\{\textbf{x}_n\}_{n=1}^N$ are independently and identically drawn (iid) from the target distribution $P_\mathcal{X}$. 

Crucially, if the samples are not iid, $\mathcal{L}(\pmb{\theta})$ will no longer be an unbiased approximator of $D_{\text{KL}}$. Then minimizing $\mathcal{L}(\pmb{\theta})$ will produce an NF model $F_{\pmb{\theta}}$ that is a biased approximation of $P_{\mathcal{X}}$ \citep{noniid_nf}, but a faithful fit to the imperfect MCMC samples, pretending they were iid. The resulting NF will hence include information on the imperfections of the MCMC chain. 

Consequently, if a problematic MCMC chain is split into several subsections, and NF models are trained on each of these subsections, then the resulting NFs will all differ from each other. Each of these NFs will then assign a low probability to samples of other sub-sections of the chain.

In brief, for a bad chain, each NF will be a bad statistical model for samples from other sections of the chain, whereas a good chain will result in NFs that agree. The new MCMC convergence diagnostic we will detail upon in the following Sec.~relies on this discrepancy. In practice, the direct comparison of an NF model to unseen samples is dependent on the exact $P_{\mathcal{X}}$ and hyperparameters of the NF architecture. To drop this dependency, a normalization via training of several flows is necessary. These details will also be discussed in the following section.  

\section{\texttt{NFD\lowercase{iag}}: MCMC Convergence using \\ Normalizing Flows} \label{subsec:test_intro}
An MCMC sampler builds up a chain of length $L$ of samples of the parameters $\Theta$\footnote{Not to be confused with the hyperparameters $\pmb{\theta}$.} generated from a posterior function $P$ conditioned on the data $\textbf{d}$ given by $P(\Theta|\textbf{d})  \equiv P$. Often one runs several such chains, and we denote the $i$th such MCMC chain as $\mathsf{S}_i = \{\Theta_{i,k}\}_{k=1}^L$, where $i \in [1,m]$ for $m$ chains. Each of these chains will separately satisfy that its number density $n_i(\Theta)$ of samples will be proportional to \emph{a} probability density $P_i$
\begin{equation}
    n_i(\Theta) \propto P_i(\Theta).
    \label{eq:2.6}
\end{equation}

Chain $\mathsf{S}_i$ is considered converged when the density of samples $P_i$ is proportional to the target density $P$ over the entire domain of $\Theta$. If the chains are not converged, then these individual densities $P_i$ will differ from each other, $P_i(\Theta) \neq P_j(\Theta), \forall i\neq j$, and this implies that they cannot be trusted to represent the target density $P(\Theta|\textbf{d})$.

Let $\mathsf{S}_1$ and $\mathsf{S}_2$ be two chains. To quantify whether they represent compatible distributions, we train an NF model $F_{\pmb\theta}$ on  $\mathsf{S}_1$ and compare it to $\mathsf{S}_2$ such that the second chain serves as validation chain. 

To do so, we evaluate the NF-based probabilities $F_{\mathsf{S}_i} = \{F_{\pmb\theta}(\Theta_{i,k})\}_{k=1}^L$ for $i \in [1,2]$ for the samples $\Theta_{i,k}$ in the two chains, and compute a distance $d(F_{\mathsf{S}_1}, F_{\mathsf{S}_2})$ between them. We detail upon this distance metric in Sec.~\ref{subsec:distance}.  This distance measures whether an NF model trained on one chain assigns a large probability to the samples of another chain, which will be the case if the chains are good, and otherwise not.

If both $\mathsf{S}_1$ and $\mathsf{S}_2$ are converged to target, then $F_{\mathsf{S}_i} \propto P$ and
\begin{equation}
    \begin{aligned}
        d(F_{\mathsf{S}_1}, F_{\mathsf{S}_2}) &= T(L), \\
    \end{aligned}
    \label{eq:2.7}
\end{equation} 
for some threshold $T(L)$ that depends on the chain length $L$ and act as $\lim_{L\rightarrow\infty}T(L) = 0$ due to Eq.~\ref{eq:2.6}. 

If either are not converged, than $F_{\mathsf{S}_1} \not\propto F_{\mathsf{S}_2}$, and $ d(F_{\mathsf{S}_1}, F_{\mathsf{S}_2}) > T(L)$ due to the biases in $F_{\pmb\theta}$ as outlined in Sec.~\ref{subsec:sens}. Thus, computing the probability evaluations $F_{\mathsf{S}_i}$ via the NF model, followed by the distance $d$ between them allows us to quantify the compatibility and convergence to target of the chains $\mathsf{S}_1$ and $\mathsf{S}_2$.

\subsection{Distance Metric for Convergence Assessment} \label{subsec:distance}
For the distance metric $d$, we adopt the Bhattacharyya distance \citep{Bhattacharyya} as implemented in OpenCV \citep{opencv}. The advantage of this distance is that it is bounded to the interval $[0,1]$, where the minimum is achieved for identical distributions. For two histograms $H_1$ and $H_2$ with $N$ bins, OpenCV implements this distance by 
\begin{equation}
    d_B(H_1, H_2) = \sqrt{1 - \frac{1}{\sqrt{\bar{H_1}\bar{H_2}N}}\sum_I\sqrt{H_1(I)\times H_2(I)}},
    \label{eq:3.1}
\end{equation}

\noindent where the index $I$ runs over the bins of the histograms and $\bar{H_1}, \bar{H_2}$ are the mean counts. We illustrate $d_B$ below in Fig.~\ref{fig:db_e}, where we plot histograms of sample densities drawn from two Gamma distributions $f(x, a)$ for $a = [2, 4]$. 

\begin{figure}[htb]
    \begin{center}
        \hspace{-0.8cm}
        \centerline{\includegraphics[width=1.1\columnwidth]{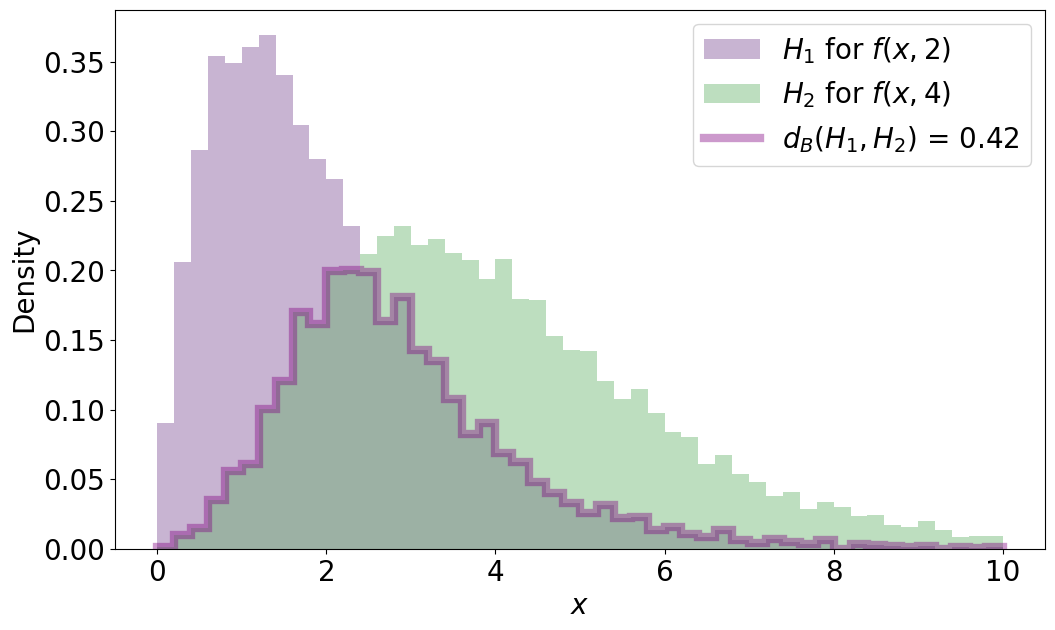}}
    \end{center}
    \vspace{-0.4cm}
    \caption{Sample density histograms $H_1$ and $H_2$ for two Gamma distributions $f(x, a)$ for $a = [2, 4]$. We outline the overlap region of the two histograms with dark purple. As the volume of the overlap region increases, $d_B$ decreases.} 
    \label{fig:db_e}
\end{figure}

As such, we compute the distance $d(F_{\mathsf{S}_1}, F_{\mathsf{S}_2}) = d_B(H_{\mathsf{S}_1}, H_{\mathsf{S}_2})$, where $H_{\mathsf{S}_i}$ are the binned probability evaluations $F_{\mathsf{S}_i}$. We discuss how $\texttt{NFDiag}$ automatically optimizes binning in Appendix \ref{sec:bhatta}. 

Both binning and NNs are dependent on some hyperparameters, which could affect $d_B$. One cannot ignore this dependency, as similar values of $d_B$ could then indicate different states of convergence across problems with different target distributions $P_{\mathcal{X}}$. To remove this dependency, we normalize $d_B$ in the following manner to reduce its sensitivity to hyperparameters of binning, the NF models and the shape of the target distribution. 

We first join the two MCMC chains $\mathsf{S}_1$ and $\mathsf{S}_2$ used in training and validation respectively. We shuffle the positions of the samples in this combined chain such to destroy the correlations between successive samples. Then, we re-separate the shuffled combined chain into two random chains $\mathsf{R}_1$ and $\mathsf{R}_2$ such that the length of $\mathsf{R}_i$ matches length of $\mathsf{S}_i$. 

Then, we train a second NF model we name $F_{\text{pair}}$ whose architecture and hyperparameters are identical to those of $F_{\pmb\theta}$. We compute a second distance $d_{B, \text{pair}} = d_B(H_{\mathsf{R}_1}, H_{\mathsf{R}_2})$ used in normalization. Then, the normalized diagnostic we call $\texttt{NFD}$ is given by
\begin{equation}
    \texttt{NFD} = \frac{1-d_B}{1-d_{B, \text{pair}}}.
    \label{eq:nfd}
\end{equation}

This way, we succeed in dropping any variation in $d_B$ due to differences in training of the NF model, binning of the histograms, and original target distribution. The resulting statistic $\texttt{NFD}$ is then such that $\texttt{NFD} \approx 1$ corresponds to total convergence and $\texttt{NFD} \approx 0$ to total lack of overlap in the number densities of the samples in the two chains. 

A minor side effect of the division in Eq.~(\ref{eq:nfd}) is that $\texttt{NFD} \gtrsim 1$ is possible when samples are iid due to variances in model fitting or $d_B$ computation. This means our convergence diagnostic can, due to noise, slightly exceed the value of unity in case of a perfect chain.

We provide a short user guide for this statistic in the following section, and demonstrate its performance in Section~\ref{sec:demos}.
\section{User Guide for \texttt{NFD\lowercase{iag}}} \label{sec:implementation}

\texttt{NFDiag} is our NF-based code that uses MCMC chains as input and computes a convergence diagnostic \texttt{NFD} for them. In this section, we detail on the code of \texttt{NFDiag}, and provide a guide of its features for users.

For the NFs, we use neural spline flows \citep{splines}: rational quadratic splines with knot points and derivatives predicted by NN. \texttt{Normflows} \citep{normflows} provides both the masked coupling and the autoregressive neural spline flows\footnote{An in-depth explanation of these can be found in Sec.~3 of \cite{flows}.}. We acknowledge that the latter would be faster to train and evaluate for probabilities, but we have used the former for faster sampling for diagnostics. The user can swap $\texttt{NFDiag}$ between the variants, and we have noticed no difference in results obtained from doing so. 

A central aspect of NN-based research is the potential of overfitting \citep{overfit}. To combat this, we utilize dropout \citep{dropout}, L2 regularization \citep{l2reg} and early stopping. A common approach is to terminate training as soon as the mean validation loss increases. However, there is evidence that this is an oversimplification of real NN training \citep{early_stopping}. 

We implement a early stopping method that assigns a penalty for increases in the loss function and $d_B$ until a budget for the penalty is exceeded. To accurately compute\footnote{Since our models are trained with mini-batches, the weights of the neural network are updated after every mini-batch. Therefore the loss computed for that mini-batch are no longer a valid statistic.} the training losses, after every epoch and freezing the NN weights, the loss in Eq.~\ref{eq:2.4} is reevaluated on the entire training chain in addition to the validation chain. Then, once the cumulative penalties exceed the budget, the training is stopped. The penalties and the budget are hyperparameters that control the trade-off between the generalization of the model and the training time and can be tuned by the user to adapt to their use case.

To run \texttt{NFDiag} for assessing the quality of their MCMC chains, the user needs to provide 
\setlist{nolistsep}
\begin{itemize}[noitemsep, leftmargin=*]
    \item[--] files for chains $\mathsf{S}_1$ and $\mathsf{S}_2$, or alternatively only $\mathsf{S}_1$ which the code then splits into $\mathsf{S}_1$ and $\mathsf{S}_2$ 
    \item[--] number of epochs to train NF models for
    \item[--] snapshot frequency, which defines the epochs where $\texttt{NFD}$ is computed and plots are produced.
\end{itemize}
$\texttt{NFDiag}$ then will automatically:
\setlist{nolistsep}
\begin{itemize}[noitemsep, leftmargin=*]
    \item[--] Assign $\mathsf{S}_1$ to training and $\mathsf{S}_2$ to validation. If only $\mathsf{S}_1$ is provided, assign the first $70\%$ to training, and rest to validation.
    \item[--] Train an NF model $F_{\pmb{\theta}}$ on $\mathsf{S}_1$ 
    \item[--] At the snapshot epochs, save a copy of the model and the probability evaluations $F_{\mathsf{S}_1}$ and $F_{\mathsf{S}_2}$.
    \item[--] Once the NF model converges or the maximum number of epochs is reached, repeat the above 3 steps, but for the pair model $F_{\text{pair}}$ which is trained instead on $\mathsf{R}_1$ as defined in Sec.~\ref{subsec:distance}.
    \item[--] Histogram the probability evaluations $F_{\mathsf{S}_i}$ and $F_{\text{pair}, \mathsf{R}_i}$ for $i \in [1,2]$ and compute $\texttt{NFD}$ for every snapshot. 
\end{itemize}

All hyperparameters of the models and NN training can also be specified by the user. We provide sensible values for these in Sec.~\ref{sec:demos} alongside our demonstrations of \texttt{NFDiag}. 
\section{Demonstrations of \texttt{NFD\lowercase{iag}}} \label{sec:demos}
In this section, we demonstrate \texttt{NFDiag} across different experiments. In Sec.~\ref{subsec:iid}, we establish that \texttt{NFD} is a universal and unbiased indicator of multivariate convergence in up to 125 dimensions. Sec.~\ref{subsec:samplers} is dedicated to applying \texttt{NFDiag} to chains from various MC samplers.

Due to runtimes, it is difficult to systematically demonstrate reliable performance of \texttt{NFDiag} in higher dimensions than 125, even though we found good performance in all higher dimensional tests we performed. We hence demonstrate in Sec.~\ref{subsec:science}, an application of $\texttt{NFDiag}$ to chains of the weak lensing power spectrum in 250 dimensions. This demonstrates the science-readiness of \texttt{NFDiag} as the weak lensing power spectrum is a common approach to inferring parameters of the current cosmological models using large scale matter distribution in the universe.

\begin{table}[htb]
    \begin{center}
    \begin{tabular}{p{0.3\columnwidth}p{0.115\columnwidth}p{0.3\columnwidth}p{0.115\columnwidth}}
        \toprule
        Hyperparameter & Value & Hyperparameter & Value\\
        \midrule
        Flow layers & 3 & Batch size & 5k \\
        NN layers & 3 & Spline knots & 20 \\
        Neurons/layer & 512 & Dropout & 0.45 \\
        Learning rate & $10^{-3}$ & Weight decay & 0.005 \\
        \bottomrule
    \end{tabular}
    \end{center}
    \caption{Hyperparameters shared between all NF models and training regimens for the experiments in Sec.~\ref{sec:demos}.}
    \label{table:hyper_shared}
\end{table}

The hyperparameters common to all NF models are displayed in Table \ref{table:hyper_shared}. We have independently verified that choosing different values will not significantly alter the results we present here and solely affect the point at which the NF models converge.

All experiments were performed on an Asus Zephyrus S GX701GWR with Intel Core i7-9750H CPU, 16.0 GB of RAM and an NVIDIA GeForce RTX 2070 GPU with 8GB of VRAM using Python 3.13.5 and PyTorch 2.7.1 with CUDA 11.8. They take around 2 to 20 minutes per repetition depending on the dimensionality.

\subsection{Universality of \textup{\texttt{NFD}} in the IID Limit} \label{subsec:iid}
The most popular subclass of MCMC algorithms are so-called `universal' samplers for which it can be shown that they correctly sample from a very wide class of target distributions. It is hence natural to ask whether \texttt{NFD} is also universally applicable to different distributions, or whether it will fail e.g. for multi-modal distributions.

For all distributions we tested, we found \texttt{NFDiag} to be of generic and stable performance. Moreover, as any unimodal and even multimodal multivariate probability distributions can be approximated as a sum of multivariate Gaussian distributions, one can expect universal stable performance of our diagnostic for any probability distribution if the \texttt{NFD} performs well for Gaussian Mixtures (GMs). To demonstrate that our diagnostic even handles multimodal distributions well, we here include into our demonstrations GM distributions in $\text{n}_{\text{dims}} = [5, 25, 75, 125]$ dimensions. For each, we create multimodal distributions consisting of $[5, 20, 100]$ Gaussian components with random mean vectors, covariance matrices and weights, and acquire $L = 100\,000$ iid samples as chains. 

\begin{figure}[htb]
    \begin{center}
        \hspace{-0.8cm}
        \centerline{\includegraphics[width=1.1\columnwidth]{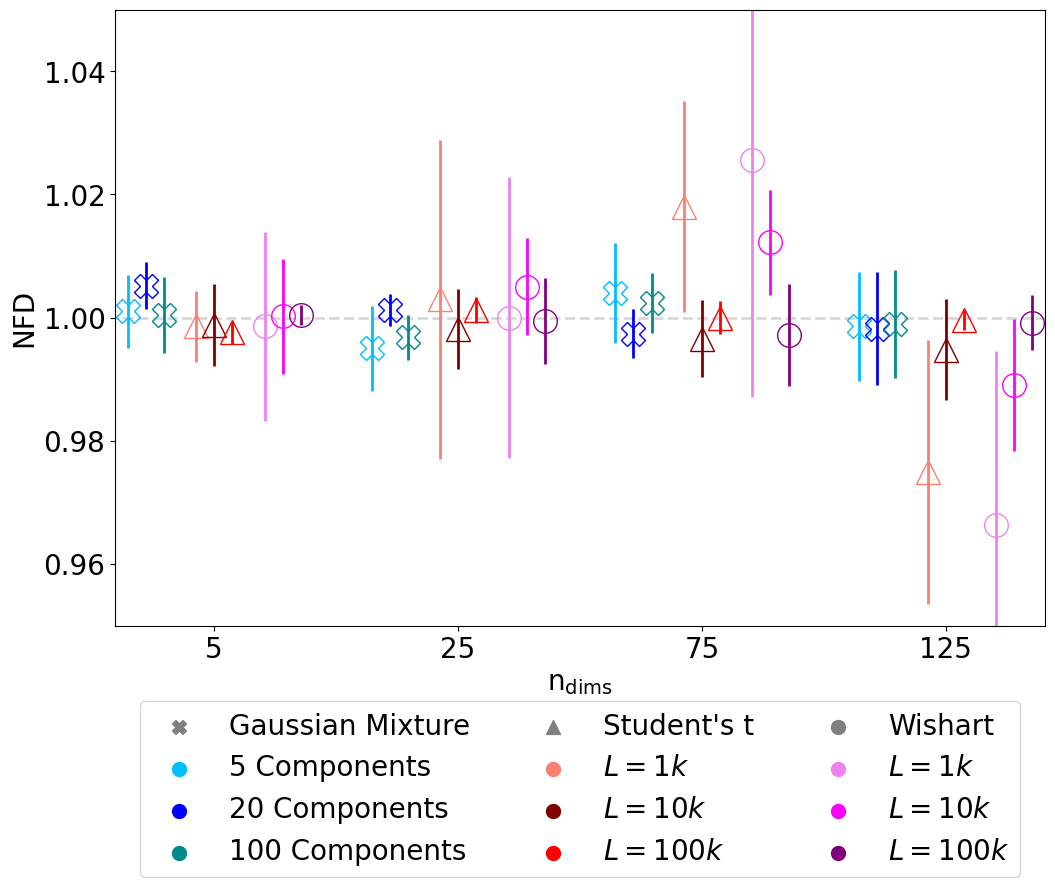}}
    \end{center}
    \vspace{-0.4cm}
    \caption{Evaluating \texttt{NFD} for iid samples in increasing dimensions, and for several distributions. The error bars are standard deviations obtained from 5 repetitions. The multivariate Student's t and matrix-variate Wishart distributions are indicated with triangles and circles respectively. With increasing $L$, the \texttt{NFD} approaches 1, indicating that it is unbiased. The Gaussian Mixture (GM) chains indicated with crosses are all of $L = 100\,000$. Since any probability distribution can be approximated through a GM, we conclude that the \texttt{NFD} is universally applicable to any target distribution as we see here that it performs reliably even for these multimodal distributions.} 
    \label{fig:iid}
\end{figure}

To establish \texttt{NFD} is unbiased, in each dimension, we generate iid chains from a multivariate Student's t, and a matrix-variate Wishart distribution with $L = [1\,000, 10\,000, 100\,000]$. Since for a set of iid samples
\begin{equation}
    \lim_{L \rightarrow \infty}\texttt{NFD} = 1,
\end{equation}
\noindent one would expect the \texttt{NFD} to asymptotically approach unity with increasing $L$ if the test is truly unbiased. The results are given in Fig.~\ref{fig:iid}, which indeed demonstrates the unbiasedness of our diagnostic across all test cases. Note that for the shorter chains, we reduce the batch size from the value in Table \ref{table:hyper_shared} to keep the number of batches constant.

\subsection{Comparisons with Other Convergence Statistics} \label{subsec:samplers}
In this section, we apply \texttt{NFDiag} to chains of samples acquired using various auto-correlated MC samplers and compare \texttt{NFD} to other statistics commonly used in assessment of convergence of MCMC chains.

First is the univariate $\hat{R}$ \citep{rhat} computed on the marginals of the multivariate chains, which aims to evaluate the stationarity of a chain via comparing the within-chain and between-chain variances. We adopt the split-$\hat{R}$ from \cite{bulk_ess}, defined as
\begin{equation}
    \hat{R} = \sqrt{\frac{L-1}{L} + \frac{1}{L}\frac{B}{W}},
\end{equation}

\noindent where $B$ and $W$ are the between- and within-chain variances of a chain of length $L$. \cite{bulk_ess} recommend $\hat{R} < 1.01$ to indicate marginal convergence. Rather than requiring each marginal to fall below this threshold, we consider our multivariate chains `converged wrt.\ $\hat{R}$' when the mean of the $\hat{R}$s is less than $1.01$.

We also compare against multivariate effective sample size (mESS) proposed by \cite{mess1}, which extends the univariate formulation. Let $\mathsf{\hat{C}}$ be the sample covariance matrix of a $d$-dimensional chain of length $L$, and $\hat{\mathsf{T}}_L$ be the Monte Carlo error estimate for this chain. Then, the mESS is given by 
\begin{equation}
    \text{mESS}= L\cdot  \left( \frac{\det{\hat{\mathsf{C}}}}{\det{\mathsf{\hat{T}}_L}} \right)^{1/d}.
\end{equation}

Crucially, we use the Monte Carlo error estimator in \cite{mess2} for greater accuracy in higher dimensions, which is a hybrid of initial sequence and batch-means estimators. Still, we notice the estimation for the correlation matrix of the mean to be dominated by noise regardless of choice of batch size. This leads to overestimates of mESS up to an order of magnitude in highly correlated chains in high number of dimensions. So, we also compute mESS using the true correlation matrix as the lowest bound. The recommendation of \cite{bulk_ess} is to aim for at least $\text{ESS} > 400$ over an ensemble of chains. In this work, we consider a stricter $\text{mESS} > 1\,000$ to indicate a chain is `converged wrt.\ mESS'.

We sample chains of a multivariate Gaussian in $\text{n}_{\text{dims}} = [5, 25, 75, 125]$ dimensions, of a fixed length $L = 100\,000$ from three samplers to compute the above statistics and \texttt{NFD} on: 

\setlist{nolistsep}
\begin{itemize}[noitemsep, leftmargin=*]
    \item[--] Metropolis-Hastings (MH) sampler with a multivariate Gaussian proposal distribution, available through \texttt{emcee} \citep{emcee, affine_invariant}. Following \cite{mh_scaling}, we set the variances of the proposal distribution to $l^2/d$ for $l = [0.1, 0.2, 0.6, 1.2, 2.38]$ where the last value provides the lowest theoretical auto-correlation in a given number of dimensions. The preceding smaller values scale the proposed step sizes down, resulting in greater auto-correlation between the samples.
    \item[--] Metropolis-adjusted Langevin algorithm (MALA) sampler \citep{mala} in PINTS \citep{Pints}. MALA proposes next states using the local gradient information, but the proposed jumps can be scaled with an $\epsilon$ hyperparameter. We run MALA with $\epsilon = [0.2, 0.3, 0.4, 0.7]$ to get chains with various auto-correlation lengths akin to the MHMC.
    \item[--] Our in-house Hamiltonian Monte Carlo (HMC) sampler \citep{almanac1, almanac2} with automated hyperparameter tuning conditioned on a target acceptance rate $\alpha$, for $\alpha = [0.3, 0.5, 0.7, 0.9]$. The higher target acceptance rates correspond to smaller proposed jump trajectories and vice versa. 
\end{itemize}

\begin{figure}[htb]
    \begin{center}
        \hspace{-0.8cm}
        \centerline{\includegraphics[width=1.1\columnwidth]{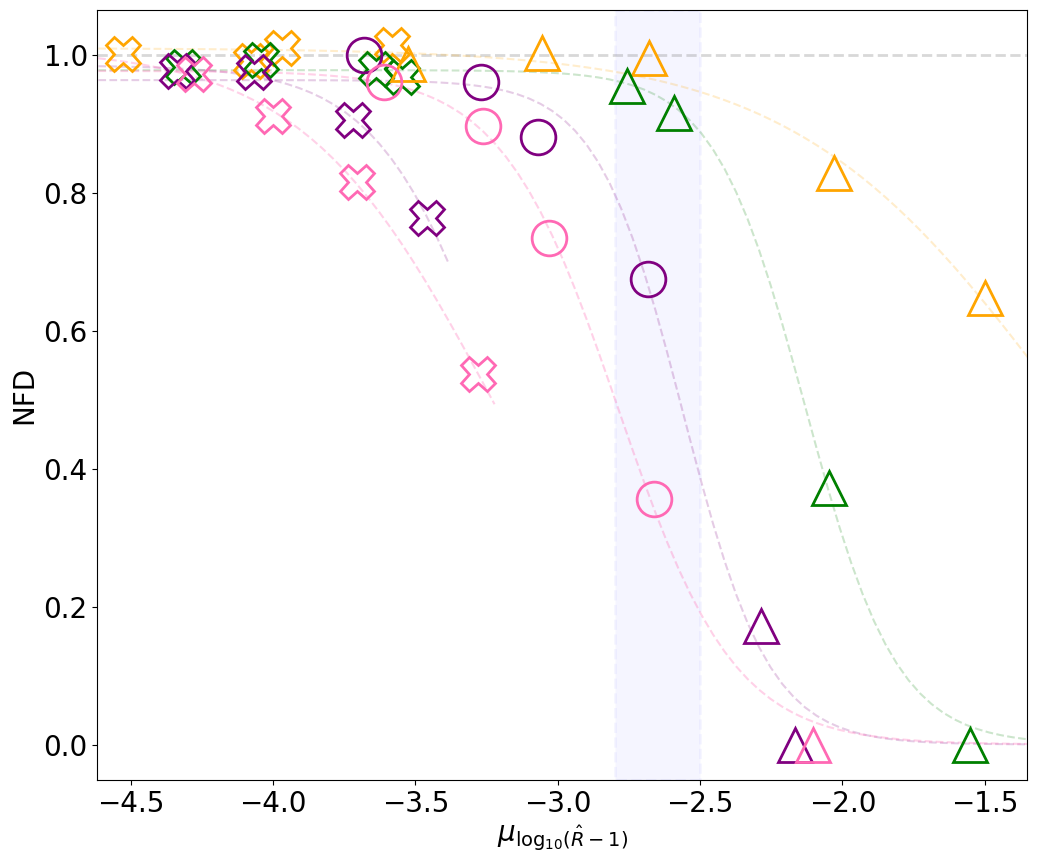}}
    \end{center}
    \vspace{-0.8cm}
    \begin{center}
        \hspace{-0.8cm}
        \centerline{\includegraphics[width=1.1\columnwidth]{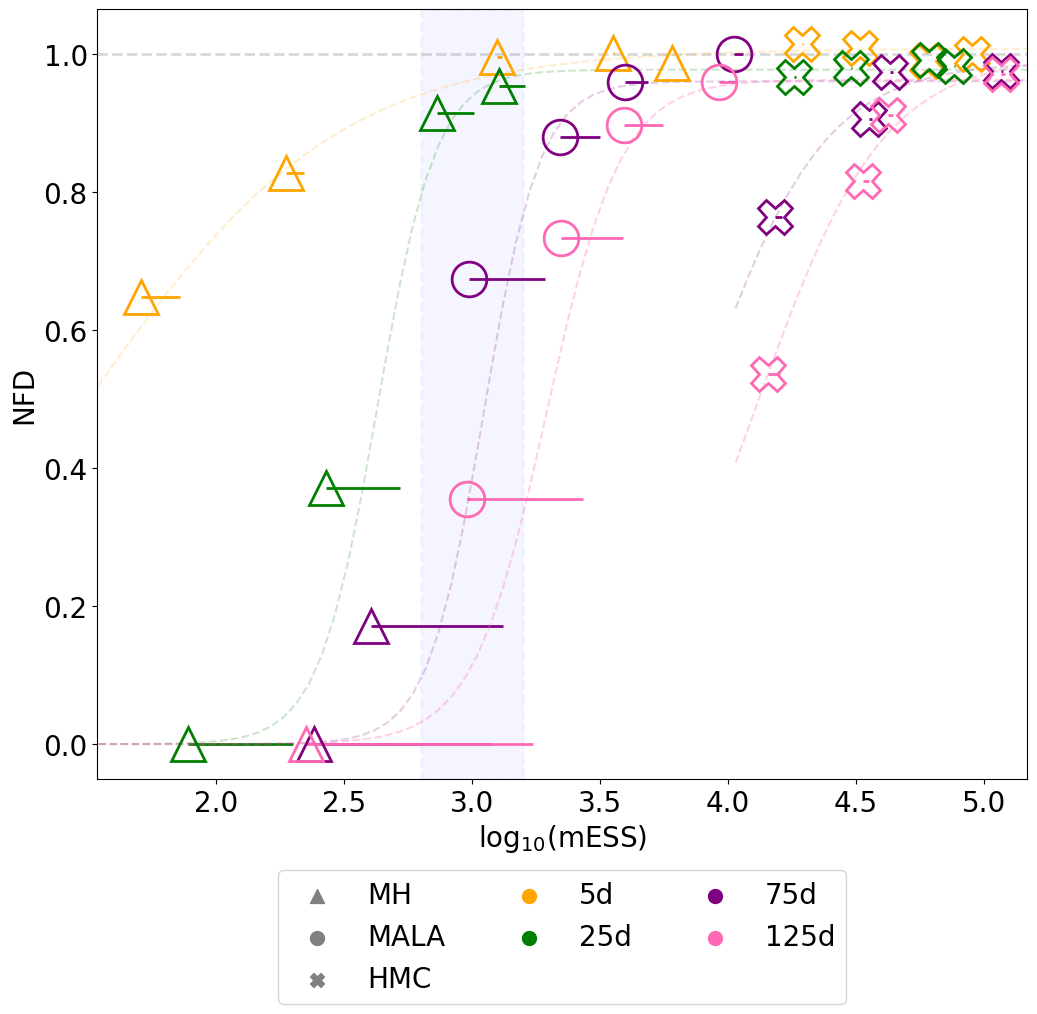}}
    \end{center}
    \vspace{-0.4cm}
    \caption{\texttt{NFD} of chains from different samplers, vs. mean $\hat{R}$ and mESS. The points indicate the mESS with the true correlation matrix, while the right-pointing error bars are the MCSE estimator with a batch size of $L^{1/3} = 47$. The dashed lines are provided to navigate the plot by connecting points of the same dimension. The vertical blue region corresponds to chains of allegedly identical convergence quality as assessed by the mESS and $\hat{R}$. Going downwards within these blue regions, one sees that these statistics become insensitive in higher dimensions, whereas \texttt{NFD} continues to resolve difference in convergence quality. As such, we see that mESS and $\hat{R}$ incorrectly indicated convergence high dimensions.}
    \label{fig:samplers}
\end{figure}

Fig.~\ref{fig:samplers} shows the comparison between \texttt{NFD}, $\hat{R}$ and mESS. Since the $\hat{R}$ and mESS adhere to the same general trend, we base our discussion mainly on the latter. Overall, for all $\text{n}_{\text{dims}}$ we have tested, the \texttt{NFD} of a`` chain increases with mESS. This indicates that the statistics share a common sensitivity to quality of the chains, and serves to increase confidence in the former's ability to assess convergence accurately. This can naturally also be extended to $\hat{R}$ comparisons.

The advantage of \texttt{NFD} comes from how it scales across $\text{n}_{\text{dims}}$ in comparison to mESS or $\hat{R}$. Generally, for a fixed value of mESS, the \texttt{NFD} decreases with $\text{n}_{\text{dims}}$, indicating that the higher dimensional chains are much further from iid (and thus convergence to target) than their lower dimensional counterparts. In turn, a dimension agnostic threshold such as $\hat{R} < 1.01$ or $\text{mESS} > 1\,000$ taken as a strong indicator of a well converged chain in a small number of dimensions might not extrapolate to the same degree of convergence in higher.

For example, all of the chains in the region around mESS $\approx 10^3$ and  $\mu_{\text{log}{10}(\hat{R}-1)} \approx -2.65$ highlighted with blue in Fig.~\ref{fig:samplers} are considered converged wrt.\ both mESS and $\hat{R}$. While the assessment wrt.\ \texttt{NFD} is in agreement for $\text{n}_{\text{dims}} = [5, 25]$, note that as the dimensionality of a distribution increases, the conditional information lost by marginalization grows monotonically. While mESS and $\hat{R}$ increasingly fail to capture this information, \texttt{NFD} is built to be highly sensitive to it: hence at $\text{n}_{\text{dims}} = [75, 125]$, the \texttt{NFD}s are significantly lower when compared to the former $\text{n}_{\text{dims}}$. In other words, where other diagnostics return over-confident convergence assessments in high dimensions, which simply originate from lost sensitivity, \texttt{NFD} continues to display dynamic behavior in large dimensions, and succeeds in assessing quality of convergence over a larger range of dimensions.

\begin{figure}[htb]
    \begin{center}
        \centerline{\includegraphics[width=0.8\columnwidth]{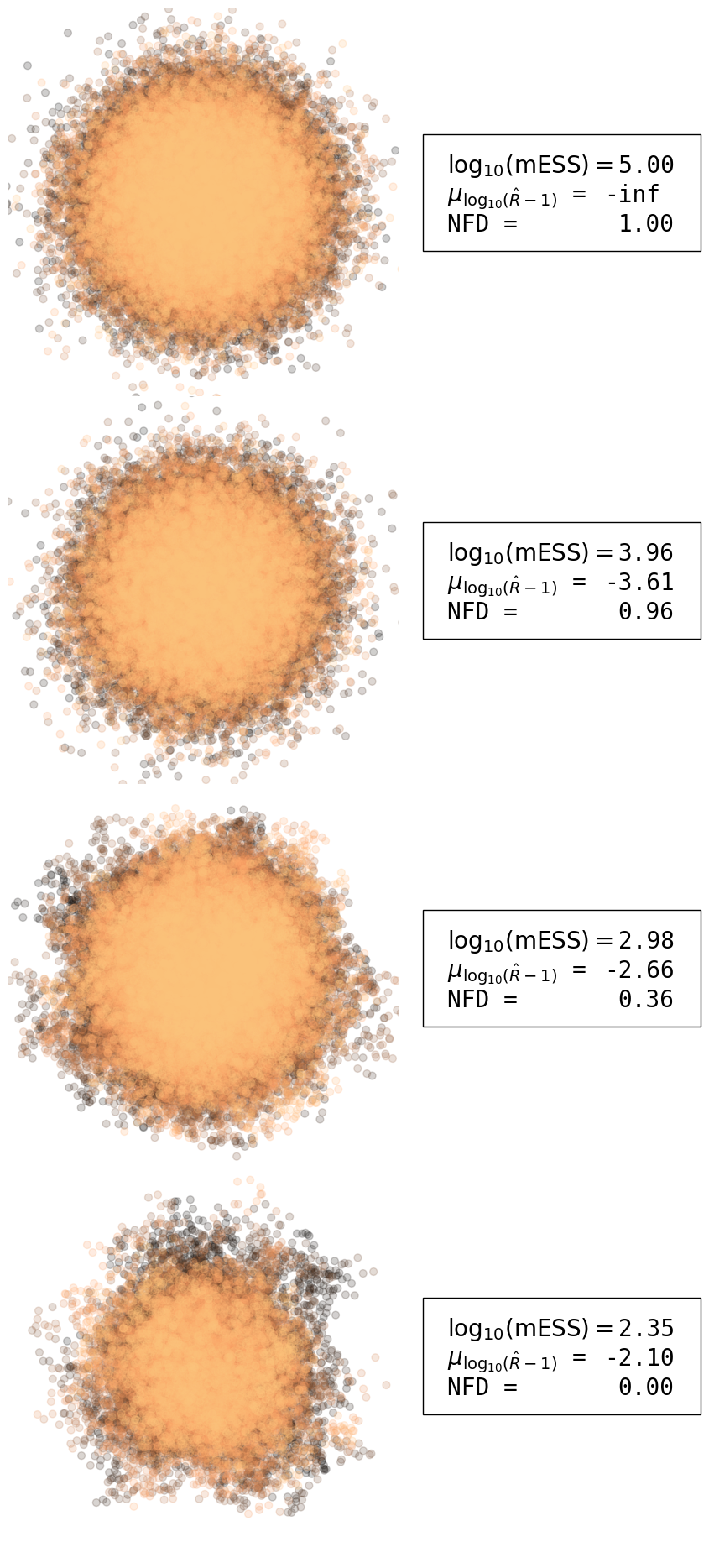}}
    \end{center}
    \vspace{-0.9cm}
    \caption{Marginal plot of, in order, iid samples, MALA chains with $\epsilon = [0.7, 0.2]$ and MH chain with $l = 2.38$, together with their mESS, $\hat{R}$ and \texttt{NFD}. The points are colored from orange to black in the sampling order of the chain. All chains are considered converged by mESS and $\hat{R}$, but visually inspecting the marginals reveals convergence problems in the latter two, which would be exacerbated in the full multivariate geometry. \texttt{NFD} is very sensitive to this, and accurately flags the chains as not converged.} 
    \label{fig:nfd_comparison}
\end{figure}

To underline visually that these significantly lower \texttt{NFD}s indeed pick up information that the other diagnostics were insensitive to, we plot in Fig.~~\ref{fig:nfd_comparison} marginals of a select 125-d chains mentioned in addition to iid samples for comparison. From top to bottom: $100\,000$ iid samples from this distribution, followed by the MALA $\epsilon = [0.7, 0.2]$, and lastly the MH chain with $l = 2.38$. All of these chains are considered converged wrt.\ mESS and $\hat{R}$, even though the human eye is able to see differences in chain quality: the bottom two chains clearly show heavier and spherically asymmetric tails, which would further be exacerbated when considering the full multivariate geometry of the chain. While this is completely missed by the mESS and $\hat{R}$, \texttt{NFD} accurately flags the problem. This visual impression aligns with the finding that with respect to \texttt{NFD}, only the $\epsilon = 0.7$ MALA chain is converged, while the others fall far away from the ideal $\texttt{NFD} \approx 1$. 

The decrease in $\texttt{NFD}$ in the transition from the best-performing MALA chains to the worst-performing HMC ones is explained by how \texttt{NFD} assesses convergence: similarity to iid samples from that same distribution. If the HMC sampler accepts a new position as the next sample, the long trajectory lengths afforded by the multi-step leapfrog integration ensures low correlation with the previous position. In comparison with algorithms that cannot execute such long jumps, chains acquired from the HMC thus have significantly higher mESS even if they accept new positions at a similar rate. However, at $\alpha = [0.3, 0.5]$, even an HMC chain will be comprised of a majority of duplicate states. As it is almost impossible for iid samples generated from a (non Dirac-delta) probability distribution to be superimposed, the \texttt{NFD}s of these chains are instead comparable to others with lower mESS. 

Taking all of the iid benchmarks and the tests on various chains from the samplers described in this section, we propose a quality cut of $\texttt{NFD} \gtrsim 0.9$ to consider an MCMC chain adequately converged wrt.\ \texttt{NFD}. Thus in practice, any value significantly smaller than this should be taken as strong indicator for a lack of multivariate convergence.

\subsection{250-dimensional Cosmology Example} \label{subsec:science}
The primary driver for the development of \texttt{NFDiag} is to provide a statistic that is sensitive and accurate in convergence assessment in high dimensional and topologically complex problems encountered in frontier scientific inference problems. In this section, we demonstrate how \texttt{NFD} can be utilized to ensure the quality of chains in weak lensing power spectrum inference. 

We replicate the analysis in \cite{almanac2}, albeit in a smaller scale: we limit ourselves to 1 tomographic bin, and generate an $N_{\text{side}} = 64$ all-sky shear map with an $\ell$-range of $[4, 128]$. The resulting joint posterior of spherical harmonic coefficients $a_{lm}$ and power spectra $\mathbf{C}_{\ell}$ is in $33\,625$ dimensions. However, in the weak lensing regime, shear is symmetric and curl-free \citep{shear1, shear2}: therefore the majority cosmological signal is embedded in the 250-dimensional combined power spectrum of $\mathbf{C}_{\mathsf{EE}, \ell}$ and $\mathbf{C}_{\mathsf{EB}, \ell}$ which we call $\mathbf{C}_{\text{signal}, \ell}$. While our in-house HMC samples from the joint posterior in $33\,625$ dimensions, we are primarily interested in the quality of the 250-dimensional $\mathbf{C}_{\text{signal}, \ell}$ samples for the scientific analysis.

This example showcases another unique strength of \texttt{NFDiag}. The probability values of samples generated by the HMC are for the joint posterior of $a_{lm}$ and $\mathbf{C}_{\ell}$, and there is no known analytic integral to marginalize out the contribution of the former to obtain probability values for just the $\mathbf{C}_{\text{signal}, \ell}$, which could be used to further inform convergence decisions. 

\texttt{NFDiag} builds precisely that distribution and can be used to assess convergence of not only full chains, but also multi-dimensional marginals of chains. This is vital in scientific inference precisely: it is common to infer nuisance parameters required by the complex Bayesian hierarchical models alongside parameters of scientific interest in scenarios where the marginal posterior distribution might not be analytically available or highly costly to evaluate numerically.

The chain of length $L = 100\,000$ we sample from this posterior is converged wrt.\ all of the following widely adopted statistics:
\setlist{nolistsep}
\begin{itemize}[noitemsep, leftmargin=*]
    \item[--] bulk ESS \citep{bulk_ess} btw. 1000-10000 per $\ell$,
    \item[--] $\hat{R}$ \citep{rhat} less than $1.01$ across all $\ell$,
    \item[--] fraction of missing information \citep{bfmi} FMI $=0.988$,
    \item[--] Hanson's diagnostic \citep{hansons} btw. $[0.95, 1.05]$ for all $\ell$. 
\end{itemize}
We provide plots for all in Appendix \ref{sec:stats}.  

\begin{figure}[htb]
    \vspace{-0.4cm}
    \begin{center}
        \centerline{\includegraphics[width=1.185\columnwidth]{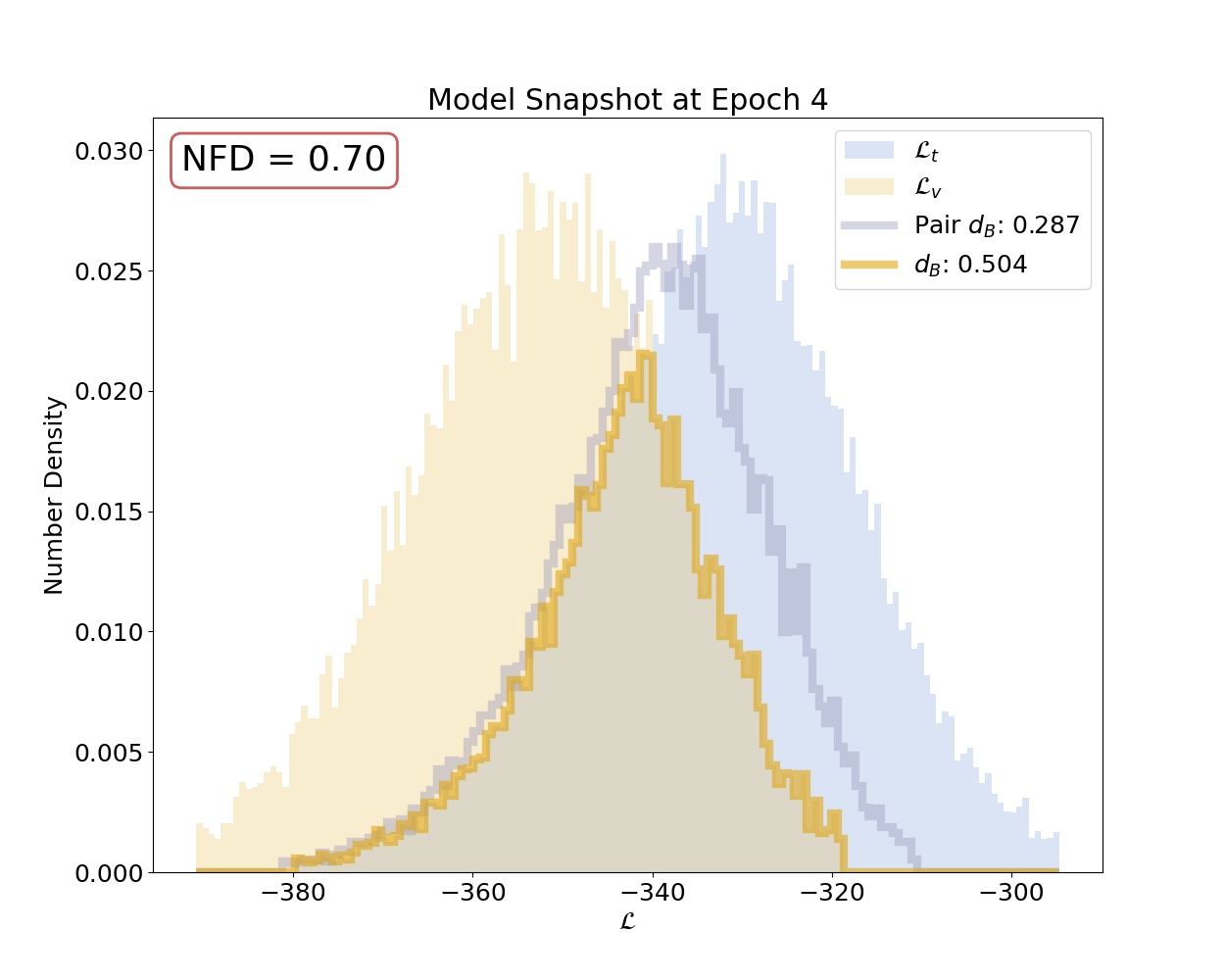}}
        \vspace{-0.45cm}
        \centerline{\includegraphics[width=1.185\columnwidth]{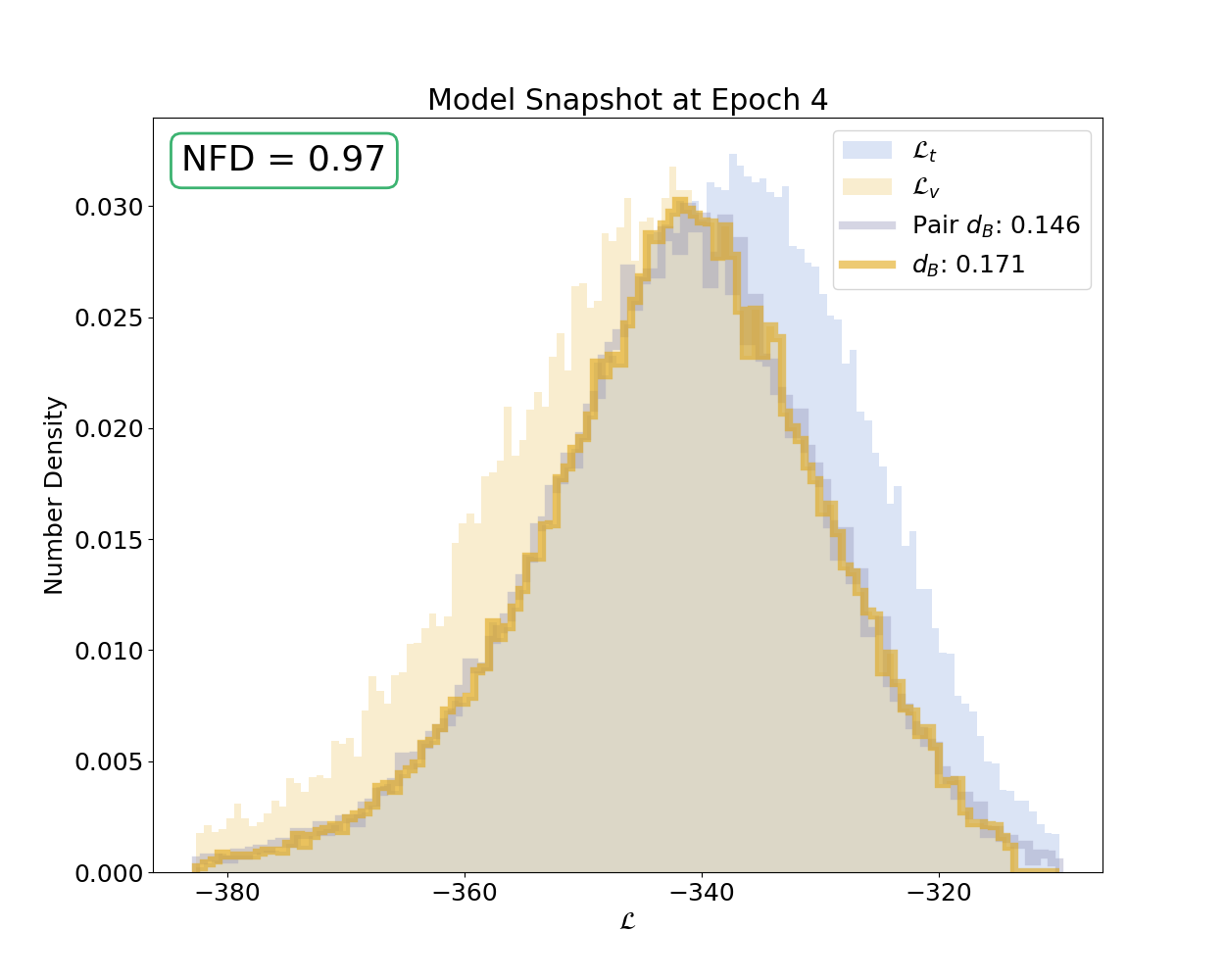}}
    \end{center}
    \vspace{-1cm}
    \caption{Histograms of the training loss $\mathcal{L}_t$ and validation loss $\mathcal{L}_v$ at Epoch $4$ for the NF model trained on (\textit{Top}) $100\,000$ $\mathbf{C}_{\text{signal}, \ell}$ samples and (\textit{Bottom}) for $100\,000$ thinned samples of $\mathbf{C}_{\text{signal}, \ell}$ from a longer chain of length $1$M from our HMC sampler.} 
    \label{fig:cosmo}
\end{figure}

The $\texttt{NFD} \approx 0.7$ is of this chain is significantly less than $\texttt{NFD} \gtrsim  0.9$ established in Sec.~\ref{subsec:samplers}. Fig.~\ref{fig:cosmo} plots the histograms of training loss $\mathcal{L}_t$ and validation loss $\mathcal{L}_v$ at model convergence and is automatically generated by \texttt{NFDiag}. The top panel corresponds to the aforementioned chain and displays the \texttt{NFD}, $d_B$ and $d_{B, \text{pair}}$. The separation between the two histograms directly corresponds to the geometric separation between the halves of the chain. In an iid sample or a well converged chain, one would expect near-perfect overlap of these histograms.

We resume sampling until we acquire a total of $1$M samples, and thin the chain to $100\,000$ samples to have an equivalent analysis ($\texttt{NFDiag}$ results with the unthinned $1$M chain are in Appendix \ref{sec:1m}). As expected, this extended and thinned chain has $\texttt{NFD} > 0.97$: similar to the multivariate Gaussian in Sec.~\ref{subsec:samplers}. The bottom panel of Fig.~\ref{fig:cosmo} displays $\mathcal{L}_t$ and $\mathcal{L}_v$ histograms for this chain, and the overlap of the two histograms has improved considerably when compared to the former chain.

The scientific inference problem in this Sec.~presents a high dimensional MCMC chain that, by various convergence tests, is sufficiently converged. Our convergence assessment with \texttt{NFDiag} utilizing the entirety of the geometric information available in the chain however indicates that the chain is yet to adequately converge. We posit such a wholistic analysis is crucial. Basing a scientific conclusion on such an MCMC chain would yield biased results, as there is no guarantee the sampler has explored the full posterior volume. Resuming the sampler for a significantly longer time generates more effective samples, and improves $\texttt{NFD}$, increasing the confidence in the convergence of the chain. 
\section{Conclusions} \label{sec:conclusions}
Accurately assessing the convergence of MCMC chains is central to avoiding biases in Bayesian inference. Only inspecting marginal distributions of samples can lead to overconfident assessment of convergence, as all information in the joint distribution is ignored. This is particularly problematic in high dimensions.

Our NFs based algorithm $\texttt{NFDiag}$ determines the convergence of high dimensional chains by modeling the posterior of samples from the chain itself. This NF model is then used to evaluate the consistency of the rest of the chain via measuring the degree with which it fails to satisfy the assumption that the samples are iid distributed. 

We demonstrate that the $\texttt{NFD}$ is a generic and unbiased metric for summarizing the degree of convergence across all dimensions. In Sec.~\ref{subsec:samplers}, we test the \texttt{NFD} on chains obtained from multiple samplers in up to $125$ dimensions.

Finally, we demonstrate $\texttt{NFDiag}$ in $250$ dimensions in a relevant Bayesian scientific inference problem, that of cosmological parameter inference from the weak lensing shear field, in Sec.~\ref{subsec:science}. We show how $\texttt{NFDiag}$ can spot chains that are not yet fully converged even when the conventional convergence statistics would indicate the contrary.

\section*{Acknowledgements}
We would like to thank Alan F. Heavens, Andrew H. Jaffe, Lorne Whiteway, Arthur Loureiro and Javier Silva Lafaurie for their help with the weak lensing inference problem in this work and their insightful discussions and comments.

\bibliographystyle{mnras}
\bibliography{main}{}

\vspace*{0cm}
\clearpage
\begin{appendices}
\section{\lowercase{$d$}$_B$ and Automated Binning} \label{sec:bhatta}
The major variable that influences the distance $d_B$ is the number of histogram bins $N$. Choosing $N$ too small would effectively smooth out over the differences in the histograms of comparison, while picking $N$ too large would cause $d_B$ to be dominated by noise.

\texttt{NFDiag} therefore computes the optimal number of bins using the Freedman-Diaconis Rule \citep{fdrule} in cases where $L \gg 1$, and with the method in \cite{brrule} when the number of samples is small. Alternatively, the user is able to specify their own desired $N$ to linearly histogram the log probability evaluations. 

In cases where strong outliers are present, if the histogram bins are prescribed between the minimal and maximal log probability values, it is highly probable that many of the bins will be under-full or simply empty due to the unrepresentative binning range. This will translate to the effective number of bins in the distance calculation to decrease and bias $d_B$ to reporting more similar distributions.

\begin{figure}[htb]
    \begin{center}
        \hspace{-0.8cm}
        \centerline{\includegraphics[width=1.1\columnwidth]{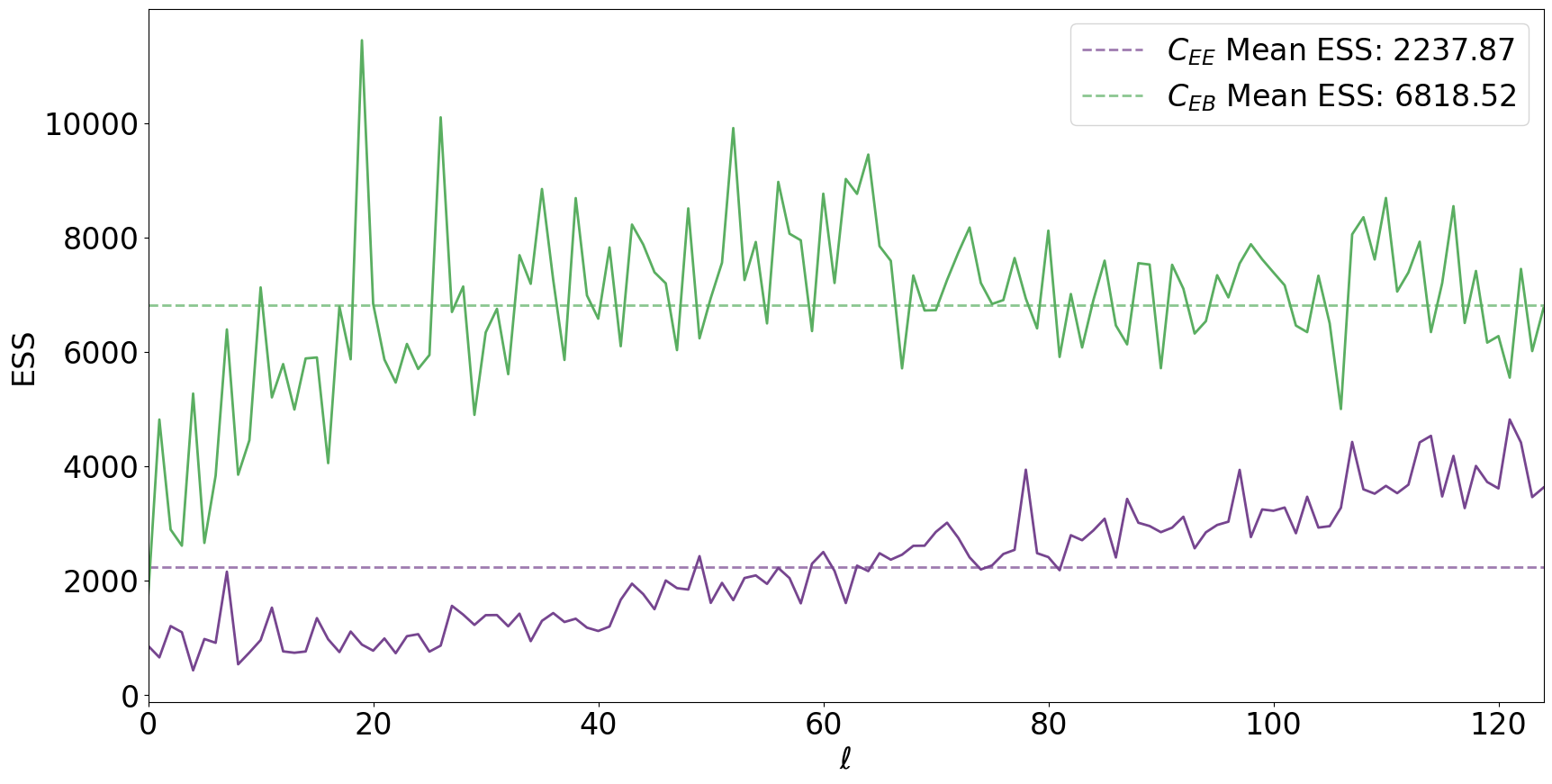}}
    \end{center}
    \vspace{-0.7cm}
    \begin{center}
        \hspace{-0.8cm}
        \centerline{\includegraphics[width=1.1\columnwidth]{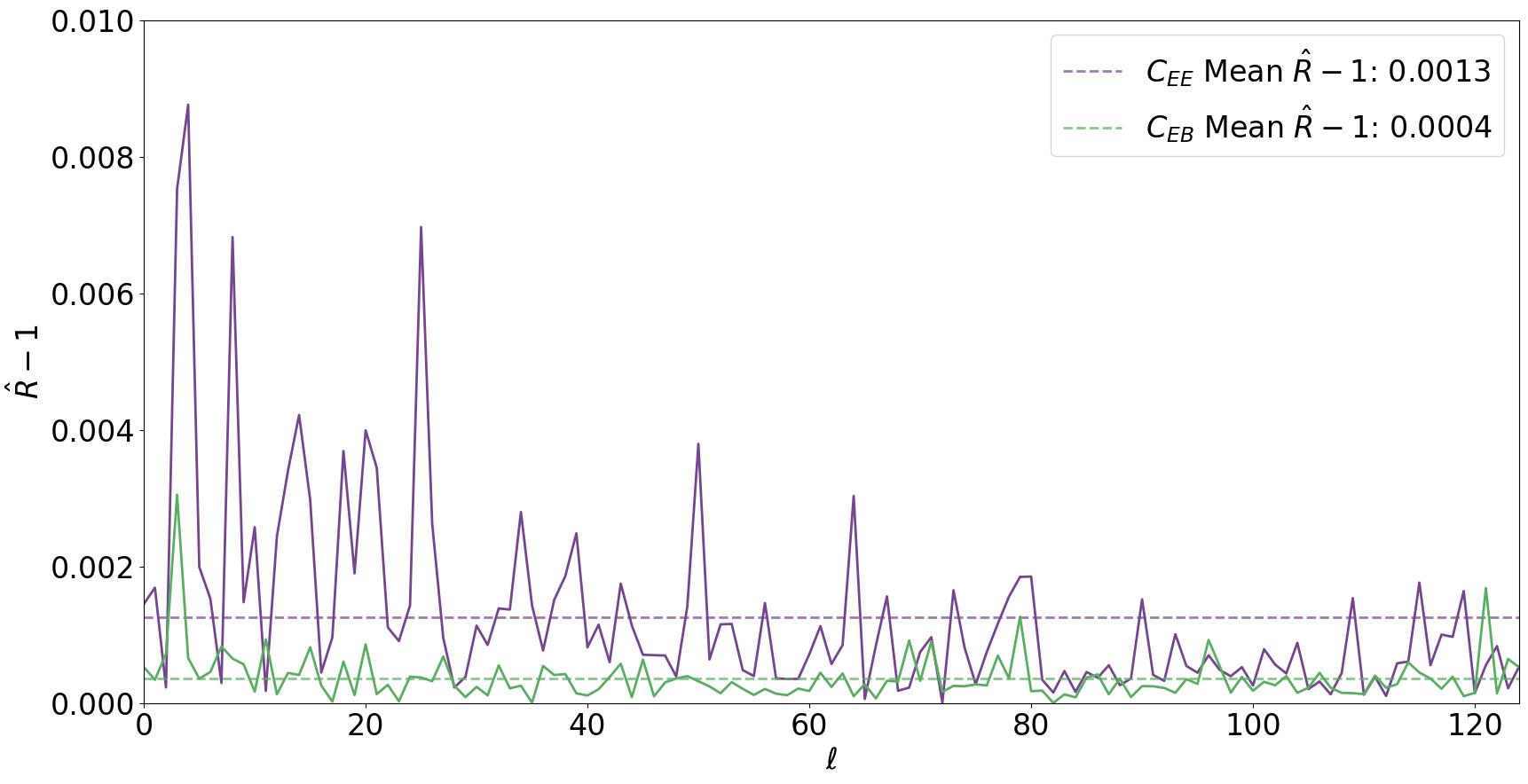}}
    \end{center}
    \vspace{-0.7cm}
    \caption{Effective sample size and $\hat{R}$ for the $100\,000$ sample chain.} 
    \label{fig:stats_1}
\end{figure}

To counteract this technicality, $\texttt{NFDiag}$ removes the bottom and top $0.5\%$ quantiles prior to histogramming. 

Additionally, $\texttt{NFDiag}$ can be asked to remove under-full bins from the tails post histogram computation, where under-full is defined as a user prescribed fraction of the highest density bin. While this additionally can help cut outlier behavior such as inadvertently included burn-in samples or spurious walkers, we caution users that removing such bins will result in $\texttt{NFDiag}$ signalling improved convergence, as the user removed a sample pathology by hand. We hence recommend users only rely on this when the source of the under-full bins can be identified with confidence. 

\section{Conventional Convergence Diagnostics} \label{sec:stats}
\begin{figure}[htb]
    \begin{center}
        \centerline{\includegraphics[width=1\columnwidth]{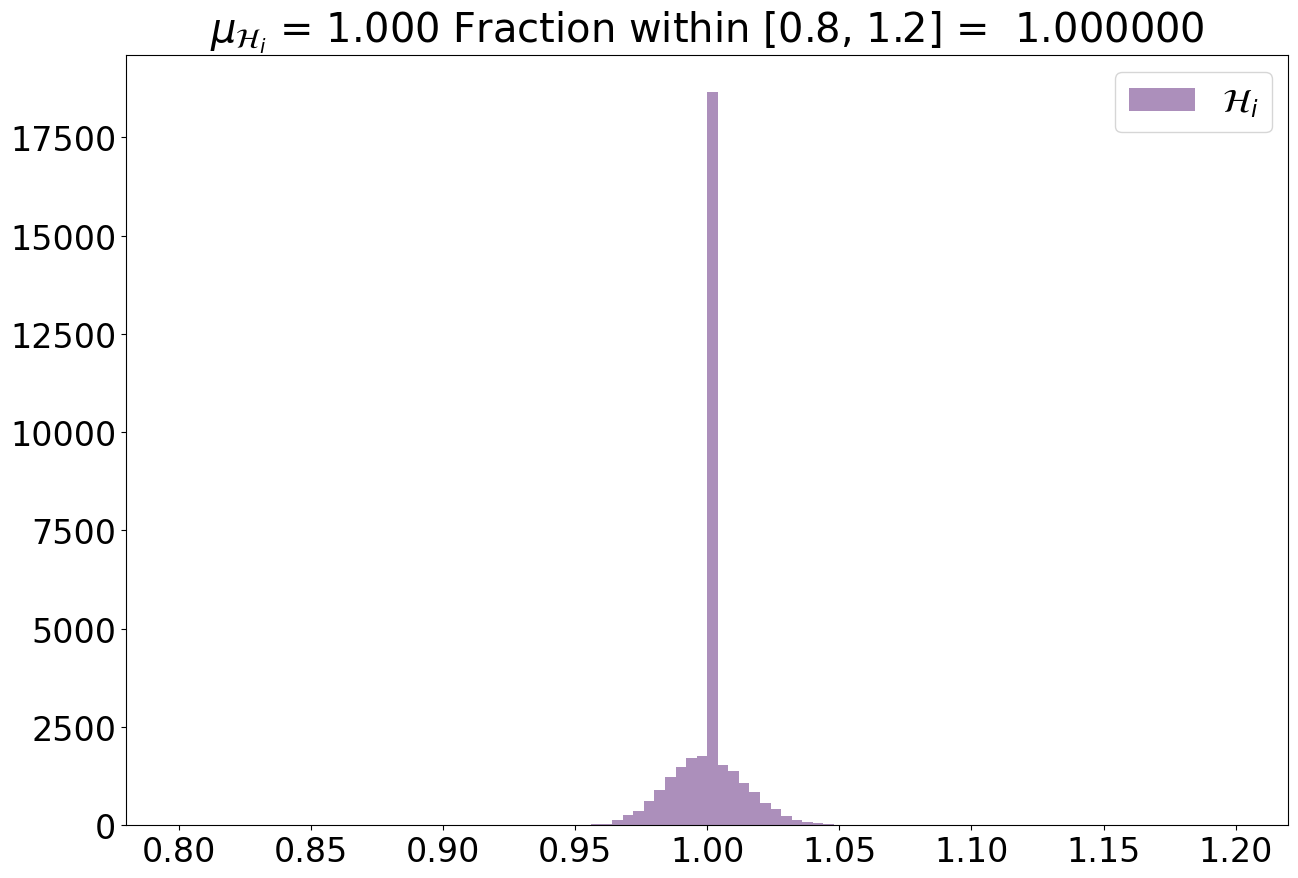}}
        \centerline{\includegraphics[width=1\columnwidth]{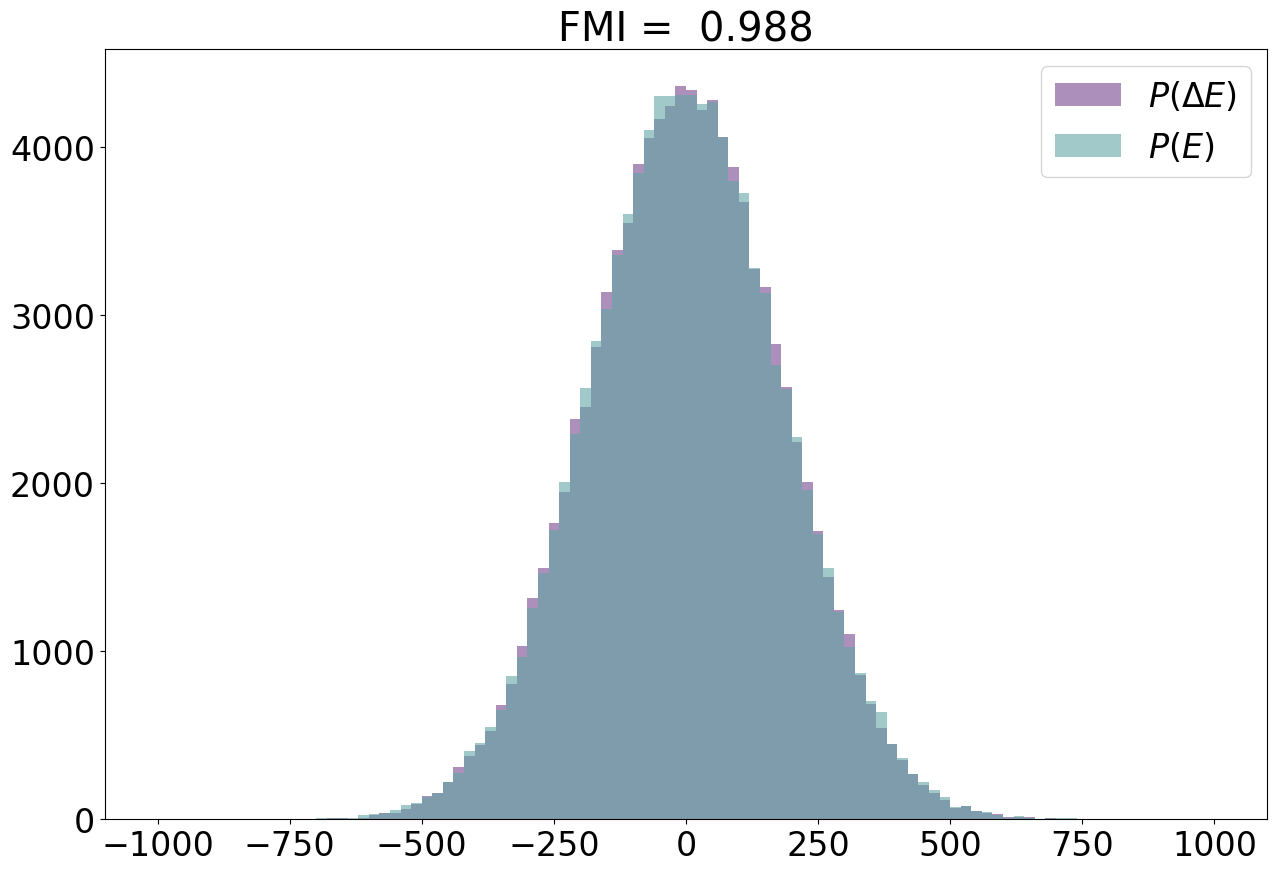}}
    \end{center}
    \vspace{-0.7cm}
    \caption{Hanson's diagnostic and FMI for the $100\,000$ sample chain.} 
    \label{fig:stats_2}
\end{figure}

In this section we provide additional plots for the convergence statistics that were quoted for the scientific inference example in Section \ref{subsec:science}. The effective sample size (ESS) \citep{ess, bulk_ess} and Gelman-Rubin $\hat{R}$ \citep{rhat} are given in Fig.~\ref{fig:stats_1}, and the Hanson's diagnostic \citep{hansons} and FMI \citep{bfmi} are given in Fig.~\ref{fig:stats_2}.

\newpage
\section{\texttt{NFDiag} of the Unthinned 1M Chain} \label{sec:1m}
This section is dedicated to evaluating the convergence of the unthinned $\mathbf{C}_{\text{signal}, \ell}$ chain of $1$M samples in addition to the thinned version presented in Section \ref{subsec:science}. Since we keep the batch size constant between the two analyses, the NF model in this section essentially updates a factor of $10$ times more per training epoch, and therefore converges rapidly and stably. The results are presented below in Fig.~\ref{fig:cosmo_1m}.

\begin{figure}[h]
    \begin{center}
        \centerline{\includegraphics[width=1.1\columnwidth]{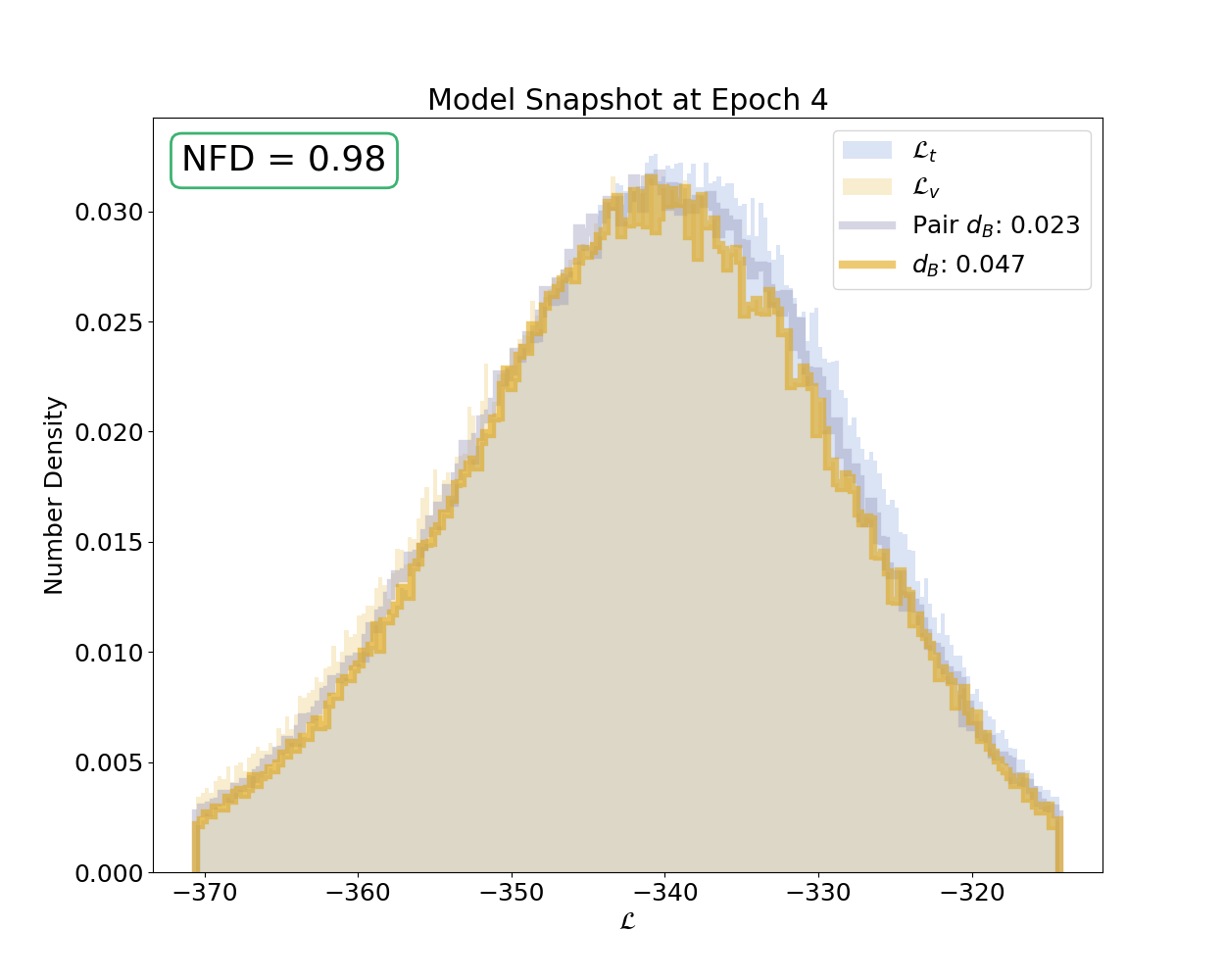}}
    \end{center}
    \vspace{-0.7cm}
    \caption{Histograms of the training loss $\mathcal{L}_t$ and validation loss $\mathcal{L}_v$ at Epoch $6$ for the NF model trained on $1$M $\mathbf{C}_{\text{signal}, \ell}$ samples from our HMC sampler.} 
    \label{fig:cosmo_1m}
\end{figure}
\clearpage

\end{appendices}

\end{document}